\documentclass[10pt,twocolumn,letterpaper]{article}

\usepackage{cvpr}              % To produce the CAMERA-READY version
\usepackage[dvipsnames]{xcolor}

\usepackage{graphicx}%
\usepackage{multirow}%
\usepackage{amsmath,amssymb,amsfonts}%
\usepackage{amsthm}%
\usepackage{mathrsfs}%
\usepackage[title]{appendix}%
\usepackage{xcolor}%
\usepackage{textcomp}%
\usepackage{manyfoot}%
\usepackage{booktabs}%
\usepackage{algorithm}%
\usepackage{algorithmicx}%
\usepackage{algpseudocode}%
\usepackage{listings}%
\usepackage{xspace}
\usepackage{makecell}

\usepackage{booktabs}
\usepackage{tabularx}
\newcolumntype{Y}{>{\centering\arraybackslash}X}
\usepackage{pifont}

\theoremstyle{thmstyleone}%
\newtheorem{theorem}{Theorem}%  meant for continuous numbers
\newtheorem{lemma}[theorem]{Lemma}

\theoremstyle{thmstyletwo}%
\newtheorem{remark}{Remark}%

\theoremstyle{thmstylethree}%

\def \snowflake {\raisebox{-.1\baselineskip}{\includegraphics[height=0.7\baselineskip]{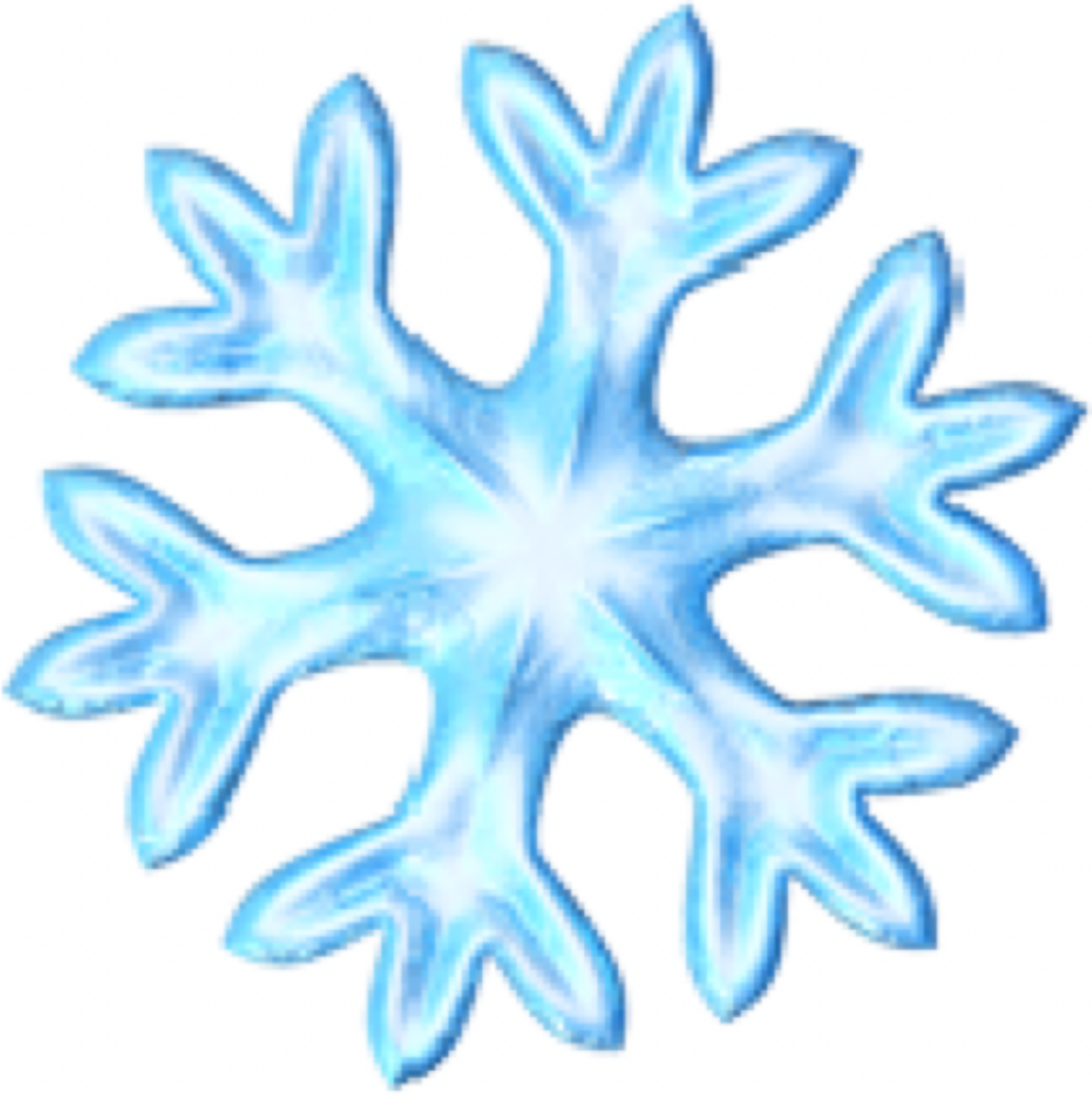}}\xspace}

\def \fire {\raisebox{-.1\baselineskip}{\includegraphics[height=0.7\baselineskip]{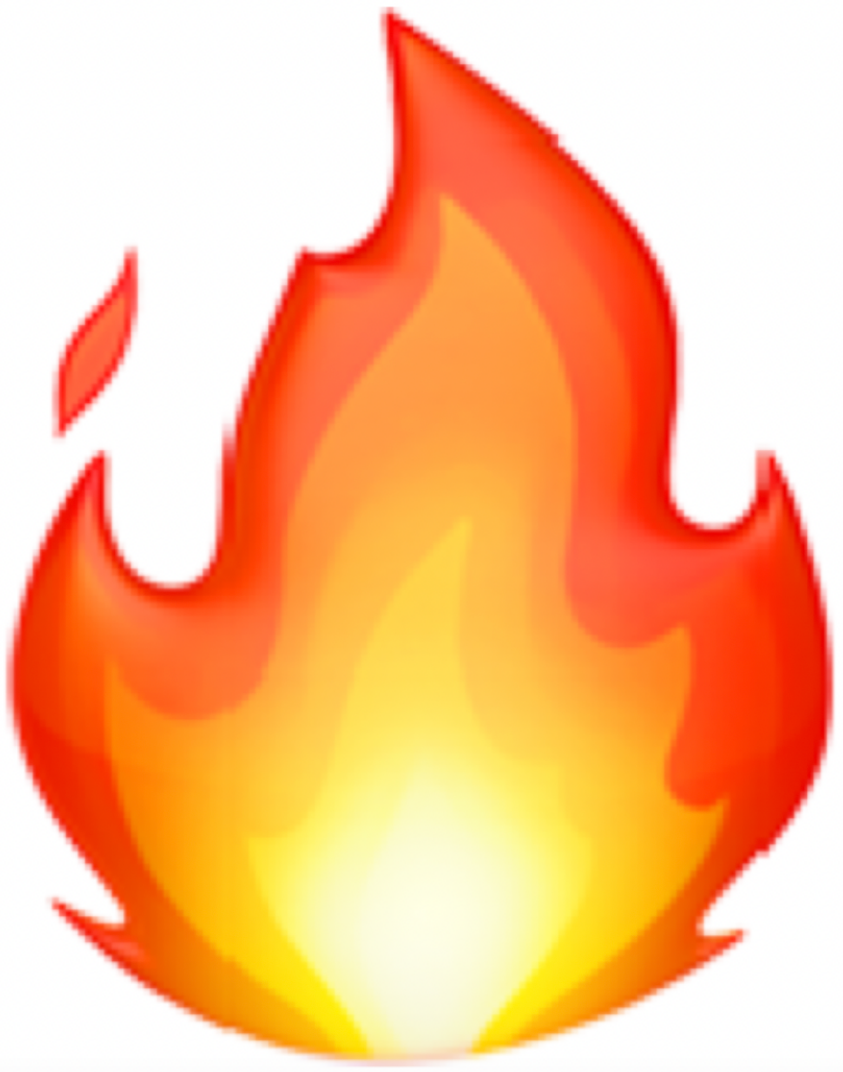}}\xspace}

\definecolor{forestgreen}{RGB}{34,139,34}
\allowdisplaybreaks

\definecolor{cvprblue}{rgb}{0.21,0.49,0.74}
\usepackage[pagebackref,breaklinks,colorlinks,citecolor=cvprblue]{hyperref}

\def\paperID{****} % *** Enter the Paper ID here
\def\confName{CVPR}
\def\confYear{2025}

\title{EZSR: Event-based Zero-Shot Recognition}

\author{Yan Yang$^{1}$ \quad Liyuan Pan$^{2}$ \quad Dongxu Li \quad Liu Liu$^{3}$ \\
$^1$BDSI, ANU \quad  $^2$BITSZ \& School of CSAT, BIT \quad  $^3$KooMap, Huawei \\
Project Page: \url{https://yan98.github.io/EZSR/} \\
{\tt\small \{yan.yang\}@anu.edu.au}
}

\begin{document}
\maketitle

\begin{abstract}
This paper studies zero-shot object recognition using event camera data. Guided by CLIP, which is pre-trained on RGB images, existing approaches achieve zero-shot object recognition by optimizing embedding similarities between event data and RGB images respectively encoded by an event encoder and the CLIP image encoder. Alternatively, several methods learn RGB frame reconstructions from event data for the CLIP image encoder. However, they often result in suboptimal zero-shot performance.

This study develops an event encoder without relying on additional reconstruction networks. We theoretically analyze the performance bottlenecks of previous approaches: 
the embedding optimization objectives are prone to suffer from the spatial sparsity of event data, causing semantic misalignments between the learned event embedding space and the CLIP text embedding space. To mitigate the issue, we explore a scalar-wise modulation strategy. Furthermore, to scale up the number of events and RGB data pairs for training, we also study a pipeline for synthesizing event data from static RGB images in mass.

Experimentally, we demonstrate an attractive scaling property in the number of parameters and synthesized data. We achieve superior zero-shot object recognition performance on extensive standard benchmark datasets, even compared with past supervised learning approaches. For example, our model with a ViT/B-16 backbone achieves 47.84\% zero-shot accuracy on the N-ImageNet dataset. 
\end{abstract}

\section{Introduction}
\label{sec:intro}
An event camera asynchronously captures pixel-wise brightness changes as an event stream, each recording the position, time, and polarity of a brightness change. Compared to traditional image sensors, event cameras not only offer benefits such as motion blur-free imaging and high temporal resolution but also consume less energy and are robust under adverse lighting conditions. The significant advantages of event cameras have led to extensive neural network applications, such as object recognition \cite{nimagnet, eventbind}, semantic segmentation \cite{eventpretraing, openess}, \etc \cite{eventdensepretraining}.

\begin{figure}[!t]
    \centering
    \includegraphics[width=0.9608\linewidth, trim={7.2pt 1pt 13.2pt 6.2pt},clip]{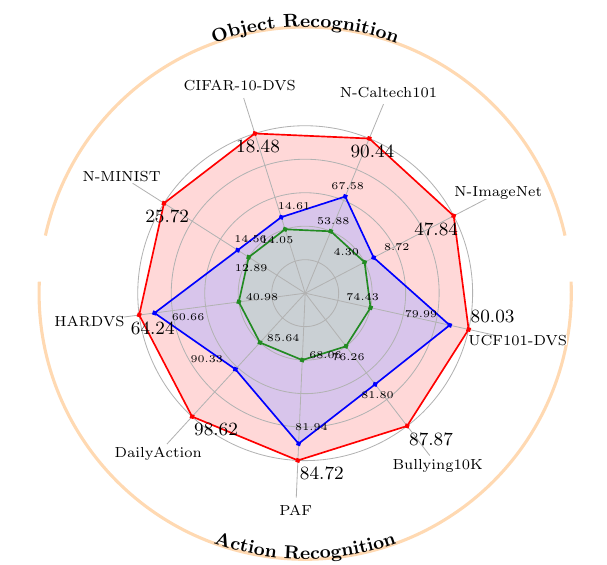}
    \vspace{-.2em}
    \caption{\it Comparison of \textcolor{red}{our} accuracies (\%) with respect to the \textcolor{blue}{second-best} and \textcolor{forestgreen}{third-best} accuracies (\%) from previous methods \cite{eventbind,eventclip,eclip} on object \cite{nimagnet,ncaltech,CIFAR-10-DVS,nmnist} and action \cite{hardvs,dailyaction,PAF, bully10k,hmdbdvs} recognition. Beside each axis, the dataset name is given. 
    }
    \label{fig:overall_comaprision}
\end{figure}

Aiming a generalized object recognition in open-world settings, this paper studies zero-shot object recognition using event camera data; in other words, our method can be tested on object classes that are not seen in the training set. Our method is contrastive learning-based, and benefited from distillation of 
pre-trained zero-shot object recognition frameworks for RGB images.
We particularly focus on CLIP \cite{clip,evaclip}, because the extraordinary success of CLIP has been demonstrated in extensive tasks.

To adapt the existing zero-shot RGB network, CLIP \cite{eclip}, to the event domain, a direct approach is to convert the event data into an event frame, and treat it as an RGB image for zero-shot object recognition \cite{eventclip,eclip}. However, event frames are typically spatially sparse, whereas RGB images densely record light intensity of a scene. The domain divergence undermines the performance of CLIP. 
In extensions, grayscale images are reconstructed from event data by learning an additional network \cite{reconstruction_clip}, enabling compatibility between the CLIP image encoder and the event data. However, it incurs extra computational costs during inference, and usually results in poor performance due to the low reconstruction quality and error accumulation between the reconstruction and recognition stages.

\begin{figure}[!t]
    \centering
    \includegraphics{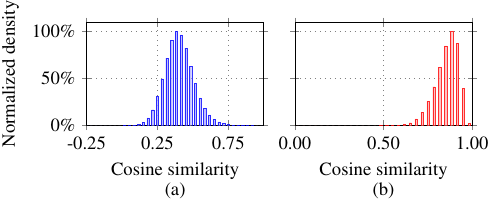}
    \vspace{-2em}
    \caption{\small \it Similarity distribution for (a) RGB and (b) event embeddings. RGB and event embeddings are extracted using a pre-trained CLIP model \cite{eva} on the validation fold of ImageNet-1K \cite{imagenet} and N-ImageNet \cite{nimagnet} datasets. Cosine similarities are computed separately among the RGB and event embeddings, and the density is normalized by the maximum value. }
    \label{fig:dist}
\end{figure}

Therefore, some works distill an event encoder from the CLIP image encoder by using paired event data and RGB images \cite{eventbind}. Afterwards, object recognition is computed from the event encoder and the CLIP text encoder. 
Contrastive learning is usually employed for learning the event encoder. However, it not only optimizes similarities between paired event and RGB data but also dissimilarities between non-paired ones. 
Due to the spatial sparsity of event data, event embeddings tend to become overly similar (\cref{fig:dist}). 
When contrasting with an image, this leads to similar similarities between the image and both paired and non-paired events, underscoring optimization attention to 
ensure the embedding discriminativeness.
Theoretically, it results in the degree of freedom for alignments between event embeddings and text embeddings from the CLIP text encoder (\cref{sec:method}). This misalignment hampers effective 
event-based zero-shot learning. 
Some works use text embeddings of class names into training, yet show  
little zero-shot performance. Moreover, due to the lack of paired event and RGB datasets, the previous approach usually performs training and testing on the same  
dataset \cite{reconstruction_clip,eventbind}, leading an 
biased 
evaluation.

This paper designs an event-based zero-shot object recognition framework for mitigating the above drawbacks.  Additionally, our method demonstrates the appealing scalability in terms of model parameters and training data. The favorable properties are credited to two key designs: scalar-wise modulation and data synthesis.

We theoretically and experimentally show that objectives involving dissimilarities optimization between non-paired event and RGB embedding lead to the semantic misalignment. Then, we propose a scalar-wise modulation strategy to directly align the event to RGB data embeddings, rather than solely relying on the contrastive learning objective. 
It compels the network to adaptively mine scalar-wise semantic alignments between RGB and event embeddings, 
dynamically directing optimization attention toward distinguishing overly similar event embeddings.
These scalar-wise alignments transfer RGB and text data alignments directly to the event and text data, and prompt the zero-shot object recognition with the CLIP text encoder.

To overcome the dataset scarcity challenge, we use synthetic data for training our network. 
Our target is zero-shot object recognition that typically relies on event data in a short duration (\eg, dozens of milliseconds) and the event data in the duration usually contains linear motion \cite{tma}. We randomly generate affine transformations and interpolate static RGB images into a sequence for event data synthesis. Compared to traditional methods that rely on video and pre-trained frame interpolation networks, our approach shows a low computation complexity, and introduces greater diversity than the same amount of video data.
Our contributions are summarized as follows:
\begin{itemize}
    \item A scalable zero-shot object recognition framework for event camera data;
    \item 
    A scalar-wise modulation strategy to promote alignments between embeddings from the event and text encoders; 
    \item A large-scale and diverse dataset of event and RGB pairs..
\end{itemize}
Experimentally, we evaluate our network on nine standard event datasets, demonstrating competitive performance even compared to the dataset-specific methods. Refer to \cref{fig:overall_comaprision} for an overall comparison. Our code and dataset will be made publicly available.

\section{Related Works}

We review the recent achievements of event camera data in zero-shot learning and relevant self-supervised learning methods. Then, we introduce the event datasets.

\begin{figure}[!t]
    \centering
    \includegraphics{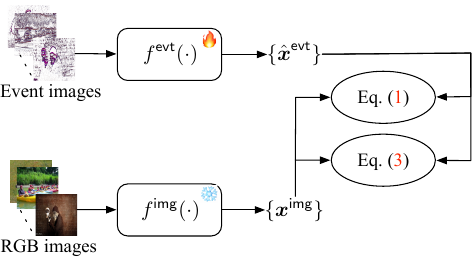}
    \vspace{-.5em}
    \caption{\it Overview of our method. Our goal is to learn an event encoder $f^{\mathsf{evt}}(\cdot)$ to replace the image encoder $f^{\mathsf{img}}(\cdot)$ from a pre-trained CLIP for allowing zero-shot object recognition with event data. Given paired event and RGB images, we respectively extract the embeddings $\{\hat{\boldsymbol{x}}^{\mathsf{evt}}\}$ and $\{\boldsymbol{x}^{\mathsf{img}}\}$ from $f^{\mathsf{evt}}(\cdot)$ and $f^{\mathsf{img}}(\cdot)$ to optimize \cref{eq:baseline} and \cref{eq:reg}. The fire (\ie, \fire) and snowflake (\ie, \snowflake) emojis respectively denote trainable and frozen components. 
     }
    \label{fig:method}
\end{figure}

\vspace{-4mm}
\paragraph{Zero-shot learning.} 
Event-based zero-shot learning is still in an early stage, with studies focused on object recognition \cite{eclip, eventbind, eventclip, reconstruction_clip} and semantic segmentation \cite{openess} using CLIP \cite{clip}. We are interested in zero-shot object recognition that can be broadly categorized into two groups:
i) reconstruction-based approach. With extra computational costs, it reconstructs grayscale images from event data to leverage pre-trained CLIP for object recognition. By exploiting the strong RGB domain prior from CLIP, the reconstruction network can be trained by using only event data \cite{reconstruction_clip}. 
Although it offers promising reconstruction quality, the zero-shot recognition performance is poor due to error accumulation;
ii) contrastive learning-based approach. Studies like \cite{eclip, eventbind} enforce embedding similarities between an event encoder and the pre-trained image encoder of CLIP, using paired event data and RGB images for training. The primary objective is instance discrimination between event and RGB images through contrastive learning.
However, as analyzed in \cref{sec:intro}, their performance is hindered, and often requires sacrificing large-scale datasets intended for evaluation to be used for training.
Other methods \cite{eventclip, eventbind} convert event data into event frames to explore the direct use of CLIP for zero-shot recognition or few-shot learning.
This paper studies a method trained on synthetic data to address their limitations and improve zero-shot object recognition.

\vspace{-4mm}
\paragraph{Self-supervised learning.} 
Similar to zero-shot learning, self-supervised learning aims to learn generic feature embeddings of event data. It employs tasks like masked image modeling \cite{mem, mae}, contrastive learning \cite{eventpretraing, mocov3}, and self-distillation \cite{eventdensepretraining, dinov2}. 
These methods improve performance on downstream tasks through transfer learning.
However, unlike zero-shot learning, they are required to learn a decoding head for recognition. This paper focuses on zero-shot object recognition on event data, enabling recognition with a simple forward pass without further training.

\vspace{-4mm}
\paragraph{Event datasets.} 
Event cameras offer high temporal resolution, reduced motion blur, and lower power consumption, making them ideal for dynamic and low-light environments. This has spurred the development of various event-based datasets for object and action recognition. 
Notable object recognition datasets include N-ImageNet \cite{nimagnet}, N-Cars \cite{ncars}, CIFAR10-DVS \cite{CIFAR-10-DVS}, N-Caltech101 \cite{ncaltech}, DVS-128-Gesture \cite{DVS128gesture}, \etc \cite{asldvs,nmnist,nsod}. Examples of the action recognition datasets are HMDB-DVS \cite{hmdbdvs}, UCF-DVS \cite{hmdbdvs}, PAF \cite{PAF}, DailyAction \cite{dailyaction}, and HARDVS \cite{hardvs}. This paper extensively evaluates our method on object recognition and action recognition datasets.

\section{Method}
\label{sec:method}
We first introduce the background of contrastive learning by establishing a baseline. Next, our framework is presented component by component, tracking the performance variations of the baseline. All training is conducted on our synthetic dataset, and the N-ImageNet, the largest event-based recognition dataset, is used for benchmarking.

\subsection{Background} 
\paragraph{Preliminary.} Contrastive learning pulls embeddings of the same instance close to each other and pushes apart embeddings of different instances. It embeds instances into a query set $\{\boldsymbol{q}\}$ and a key set $\{\boldsymbol{k}\}$. For each query embedding $\boldsymbol{q}$, there is a matching key embedding $\boldsymbol{k}_{+}$ and a set of non-matching key embeddings $\{\boldsymbol{k}_{-}\}$. Then, an embedding space is optimized to ensure that $\boldsymbol{q}$ is close to $\boldsymbol{k}_{+}$ and distant from $\{\boldsymbol{k}_{-}\}$ through minimizing an InfoNCE loss \cite{infonce}. 
The InfoNCE loss for a query embedding $\boldsymbol{q}$ is given by 
\begin{align}
\mathcal{L}_\text{nce}(\boldsymbol{q},\{\boldsymbol{k}\}) = - \log \frac{\exp(\boldsymbol{q} \cdot \boldsymbol{k}_{+} / \tau )}{\exp(\boldsymbol{q} \cdot \boldsymbol{k}_{+} / \tau ) + \sum\limits_{\boldsymbol{k}_{-}} \exp(\boldsymbol{q} \cdot \boldsymbol{k}_{-} / \tau )} \ , \nonumber
\end{align}
where $\tau$ is a temperature parameter to control the distribution sharpness. Usually, the query $\boldsymbol{q}$ and key $\boldsymbol{k}$ embeddings are $\ell_{2}$-normalized, and the dot product $\cdot$ between them calculates their cosine similarity.

\subsection{Baseline}
In the context of learning an event encoder using a pre-trained CLIP image encoder, paired event data and RGB images are treated as the same instances. The collections of event and RGB data are respectively encoded into two embedding sets: $\{\hat{\boldsymbol{x}}^{\mathsf{evt}}\}$ for events and $\{\boldsymbol{x}^{\mathsf{img}}\}$ for images, with an event encoder $f^{\mathsf{evt}}(\cdot)$ and a pre-trained CLIP image encoder  $f^{\mathsf{img}}(\cdot)$. The embeddings $\{\hat{\boldsymbol{x}}^{\mathsf{evt}}\}$ and $\{\boldsymbol{x}^{\mathsf{img}}\}$ are mutually served as query set and key sets for optimizing a symmetrized InfoNCE loss\footnote{For brevity, we take the symmetrized InfoNCE as an example. Refer to our supplementary materials for analysis of other objective functions.},
\begin{align}
    \mathcal{L}&_{\text{baseline}} = \frac{1}{\lvert \{\hat{\boldsymbol{x}}^{\mathsf{evt}}\} \rvert }\sum_{\hat{\boldsymbol{x}}^{\mathsf{evt}} \in \{\hat{\boldsymbol{x}}^{\mathsf{evt}}\}}\mathcal{L}_\text{nce}(\hat{\boldsymbol{x}}^{\mathsf{evt}}, \{\boldsymbol{x}^{\mathsf{img}}\})  \nonumber \\
    & \ + \frac{1}{\lvert \{\boldsymbol{x}^{\mathsf{img}}\} \rvert }\sum_{\boldsymbol{x}^{\mathsf{img}} \in \{\boldsymbol{x}^{\mathsf{img}}\}}\mathcal{L}_\text{nce}(\boldsymbol{x}^{\mathsf{img}}, \{\hat{\boldsymbol{x}}^{\mathsf{evt}}\}) \ , \label{eq:baseline}
\end{align}
where $\lvert \{{\hat{\boldsymbol{x}}^{\mathsf{evt}}} \} \rvert$ and $\lvert \{{\boldsymbol{x}^{\mathsf{img}}}\} \rvert$ respectively denote the sizes of event and image embedding sets. During training, the image encoder $f^{\mathsf{img}}(\cdot)$ is frozen, maintaining the CLIP embedding space structure for directly using the pre-trained CLIP text encoder $f^{\mathsf{txt}}(\cdot)$ in zero-shot object recognition. 

Though the baseline approach can effectively minimize \cref{eq:baseline} during training, it fails to find an accurate zero-shot object recognition performance, as seen in \cref{tab:method}. The low performance, 9.57\%, suggests a divergence between the optimization and evaluation objectives.

We note the divergence is raised by the degree of freedom associated with the embedding dimension and optimization objective. Let $\boldsymbol{x}^{\mathsf{img}}_{+}$ and $\boldsymbol{x}^{\mathsf{txt}}_{+}$ embedded respectively by the pre-trained $f^{\mathsf{img}}(\cdot)$ and $f^{\mathsf{txt}}(\cdot)$ are matching RGB image and text descriptions, which are also matching with $\hat{\boldsymbol{x}}^{\mathsf{evt}}$,  and $\hat{\boldsymbol{x}}^{\mathsf{evt}}_{-}$ and $\boldsymbol{x}^{\mathsf{img}}_{-}$ be the non-matching event and image embedding.  

\vspace{-.5em}
\begin{lemma}
When \cref{eq:baseline} is effectively minimized,  $\boldsymbol{x}^{\mathsf{img}}_{+} \cdot \boldsymbol{x}^{\mathsf{txt}}_{+} > \boldsymbol{x}^{\mathsf{img}}_{-} \cdot \boldsymbol{x}^{\mathsf{txt}}_{+}$ does not imply  $\hat{\boldsymbol{x}}^{\mathsf{evt}} \cdot \boldsymbol{x}^{\mathsf{txt}}_{+} > \hat{\boldsymbol{x}}^{\mathsf{evt}}_{-} \cdot  \boldsymbol{x}^{\mathsf{txt}}_{+}$, due to degree of freedom in the embedding space. 
Refer to our supplementary materials for proof. 
\label{lemma:degree}
\end{lemma}

\begin{table}[!t]
    \centering
    \caption{\it Zero-shot object recognition accuracy (\%) on the N-ImageNet dataset \cite{nimagnet} by combining our model components.}
    \vspace{-.5em}
    \begin{tabularx}{\linewidth}{cccY}
    \toprule
     $\mathcal{L}_{\text{baseline}}$ & \makecell[c]{\textbf{Remark 1}}  & \makecell[c]{Scalar-wise\\Modulation} & Accuracy  \\
    \midrule
     \ding{51}  & \ding{55} & \ding{55} & 9.57 \\
     \ding{51}  & \ding{51} & \ding{55} & 43.48 \\ 
     \ding{51}  & \ding{55} & \ding{51} & 47.84\\
     \ding{55}  &  \ding{55} & \ding{51} & 47.80\\  
     \ding{55}  &  \ding{51} & \ding{51} & 48.63\\ 
     \ding{51}  & \ding{51} & \ding{51} & 48.86 \\
    \bottomrule
    \end{tabularx}
    \label{tab:method}
\end{table}

\vspace{-1em}
\begin{remark}
The misalignment between event and text embeddings can be mitigated if a proper reference dataset is provided.  According to \cref{eq:baseline}, we know the event embedding is pulled to semantic similar image embeddings, while the image embedding is already aligned with the text embedding in the CLIP for zero-shot object recognition. Thus, we translate $\hat{\boldsymbol{x}}^{\mathsf{evt}}$ by using a pool of pre-embeded image embeddings $\{\boldsymbol{x}^{\mathsf{img}}\}$ to align with the CLIP text embedding. Let $\{k\}$ be the indices set of k-NN embedding from $\{\boldsymbol{x}^{\mathsf{img}}\}$, using cosine similarities as the distance measure. The translated event embedding $\tilde{\boldsymbol{x}}^{\mathsf{evt}}$ is 
\begin{align}
    \tilde{\boldsymbol{x}}^{\mathsf{evt}} = \sum_{k \in \{k\}} \frac{\hat{\boldsymbol{x}}^{\mathsf{evt}} \cdot \boldsymbol{x}^{\mathsf{img}}_{k} + 1}{\sum_{k' \in \{k\}} (\hat{\boldsymbol{x}}^{\mathsf{evt}} \cdot \boldsymbol{x}^{\mathsf{img}}_{k'} + 1)}\boldsymbol{x}^{\mathsf{img}} \ . \label{eq:interplote}
\end{align}
With the translated event embedding $\tilde{\boldsymbol{x}}^{\mathsf{evt}}$, the zero-shot object recognition performance can be improved from 9.57\% to 43.48\%. However, we do not focus on the strategies, as the performance is usually reference dataset dependent. For readers' interest, we present the ablations with it in \cref{tab:method}.
\end{remark}

\subsection{Scalar-wise Modulation}
One may note that there is an exemption in Lem.~\ref{lemma:degree}. If $\hat{\boldsymbol{x}}^{\mathsf{evt}} \cdot \boldsymbol{x}^{\mathsf{img}}_{+} = 1$, this would imply perfect alignment between $\hat{\boldsymbol{x}}^{\mathsf{evt}}$ and $\boldsymbol{x}^{\mathsf{img}}_{+}$. In this case, $\hat{\boldsymbol{x}}^{\mathsf{evt}}$ would inherently satisfy $\hat{\boldsymbol{x}}^{\mathsf{evt}} \cdot \boldsymbol{x}^{\mathsf{txt}}_{+} > \hat{\boldsymbol{x}}^{\mathsf{evt}}_{-} \cdot \boldsymbol{x}^{\mathsf{txt}}_{+}$ if $\boldsymbol{x}^{\mathsf{img}}_{+} \cdot \boldsymbol{x}^{\mathsf{txt}}_{+} > \boldsymbol{x}^{\mathsf{img}}_{-} \cdot \boldsymbol{x}^{\mathsf{txt}}_{+}$. 

Inspired by the observation, rather than simply enforcing the embedding similarities between $\hat{\boldsymbol{x}}^{\mathsf{evt}}$ and $\boldsymbol{x}^{\mathsf{img}}_{+}$, we introduce a scalar-wise modulation to enforce the scalar-wise alignments between $\hat{\boldsymbol{x}}^{\mathsf{evt}}$ and $\boldsymbol{x}^{\mathsf{img}}_{+}$. As noted in \cref{sec:intro}, event embeddings $\{\hat{\boldsymbol{x}}^{\mathsf{evt}}\}$ tend to be overly similar to one another, warranting greater optimization focus. 
To address it, our approach is adaptive, constraining the degrees of freedom by ensuring that each scalar of the event embedding and the image embedding are aligned, therefore enhancing the embedding alignment.

We achieve it by adding a mean squared error
to the optimization objective that adaptively penalizes the scalar-wise deviations in the embeddings. The loss is given by 
\begin{align}
\mathcal{L}_{\text{mod}} &= \sum_{\lambda_{\hat{\boldsymbol{x}}^{\mathsf{evt}}}, \hat{\boldsymbol{x}}^{\mathsf{evt}}, \boldsymbol{x}^{\mathsf{img}}_{+}}  \lambda_{\hat{\boldsymbol{x}}^{\mathsf{evt}}} \lVert \hat{\boldsymbol{x}}^{\mathsf{evt}} - \boldsymbol{x}^{\mathsf{img}}_{+} \rVert^{2} \ , \label{eq:reg} \\
\lambda_{\hat{\boldsymbol{x}}^{\mathsf{evt}}} &=  \mathcal{N} \big( 1 - \frac{\lambda_{\hat{\boldsymbol{x}}^{\mathsf{evt}}}^{\mathsf{unf}} - \min(\{\lambda_{\hat{\boldsymbol{x}}^{\mathsf{evt}}}^{\mathsf{unf}}\})}{\max(\{\lambda_{\hat{\boldsymbol{x}}^{\mathsf{evt}}}^{\mathsf{unf}}\}) - \min(\{\lambda_{\hat{\boldsymbol{x}}^{\mathsf{evt}}}^{\mathsf{unf}}\})}\big) \ , \\
\lambda_{\hat{\boldsymbol{x}}^{\mathsf{evt}}}^{\mathsf{unf}} &= \frac{\sum_{\hat{\boldsymbol{x}}^{\mathsf{evt}}_{-}} (\hat{\boldsymbol{x}}^{\mathsf{evt}} \cdot \hat{\boldsymbol{x}}^{\mathsf{evt}}_{-} + 1)}{\sum_{\hat{\boldsymbol{x}}^{\mathsf{evt'}}} \sum_{\hat{\boldsymbol{x}}^{\mathsf{evt'}}_{-}} (\hat{\boldsymbol{x}}^{\mathsf{evt'}} \cdot \hat{\boldsymbol{x}}^{\mathsf{evt'}}_{-} + 1)} \ ,
\end{align}
where $\mathcal{N}(\cdot)$ is a Gaussian distribution function, and the subscript `$-$' consistently to denote negative matching embedding (\ie, another embedding in the batch $\{\hat{\boldsymbol{x}}^{\mathsf{evt}}\}$).
$\lVert \cdot \rVert$ is the Frobenius norm that reduces the vector to a scalar. The input to the Gaussian function falls within the range from 0 to 1, ensuring that $\lambda_{\hat{\boldsymbol{x}}^{\mathsf{evt}}}$ remains non-negative.
By framing $\mathcal{L}_{\text{baseline}}$ as a regularization term to maintain embedding discriminativeness, the overall objective is
\begin{align}
\mathcal{L}_{\text{total}} = \mathcal{L}_{\text{baseline}} + \mathcal{L}_{\text{mod}} \ .
\end{align}
With the modulation, the degrees of freedom are constrained, taming the event embeddings to effectively align with the text embeddings for improving the zero-shot performance.

We present the detailed combinations of using $\mathcal{L}_{\text{mod}}$ in \cref{tab:method}. It consistently improves the performance by enforcing a stronger alignment constraint. For example, 47.84\% accuracy is achieved with our scalar-wise modulation.

\subsection{Data Synthesis}
\label{sec:datasynthesis}
\begin{figure}[!t]
    \centering
    \includegraphics{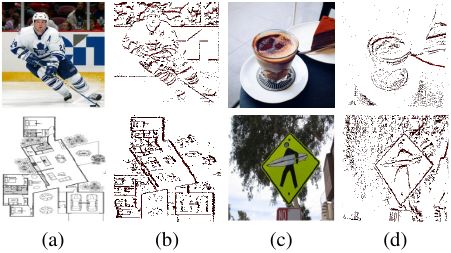}
    \vspace{-.5em}
    \caption{\it Samples of synthetic event data. (a)/(c) are RGB images. (b)/(d) are event frames, where red and blue indicate positive and negative events, respectively.}
    \label{fig:samples}
\end{figure}

Due to the lack of paired event data and RGB images for training, we generate synthetic event data. In object recognition tasks using event datasets, events captured over a short duration are typically used, emphasizing semantic information rather than motion information. Since events within a short duration often exhibit linear motion \cite{tma}, we simulate random linear motion on a static RGB image. The RGB image is warped to create a video, and the video is subsequently transformed into event data using V2E \cite{v2e}. Three motion patterns are explored, as described below:
\begin{itemize}
    \item Translation. Random horizontal and vertical maximum displacements are first sampled, and then the static image is displaced linearly to meet the maximum displacement. The translation simulates the effect of the image moving in a straight line across the frame.
    \item Scaling. The image is scaled up or down linearly, simulating zoom-in and zoom-out effects. It involves increasing or decreasing the size of the image uniformly from the center, creating the appearance of the camera moving closer to or farther away from the subject.
    \item Rotation. The image is rotated around its center by a random angle within a specified range, to simulate the effect of the image spinning in place, mimicking a rotational movement of the camera or the object within the scene.
\end{itemize}
Samples of the synthetic event data are provided in \cref{fig:samples}, and we follow \cite{eventpretraing} for visualizations. When using the training set of the N-ImageNet dataset \cite{nimagnet} for training, the 
performance is 54.36\%.
By training on the out-of-domain synthetic dataset, we can achieve competitive zero-shot object recognition performance (47.84\% vs. 54.36\%) on the N-ImageNet dataset \cite{nimagnet}.

\section{Experiments}

\subsection{Set-ups}

\paragraph{Training datasets.} To train our methods, we create a synthetic event dataset by curating RGB data from multiple sources, following the pipeline described in \cref{sec:datasynthesis}. In total, there are around 180M RGB images. Refer to the supplementary materials for details.

\vspace{-4mm}
\paragraph{Downstream datasets.} Our method is evaluated on the object and action recognition tasks. For object recognition, N-ImageNet \cite{nimagnet}, N-Caltech101 \cite{ncaltech}, CIFAR-10-DVS \cite{CIFAR-10-DVS}, and N-MINIST \cite{nmnist} datasets are explored. To benchmark our performance on the action recognition tasks, we explore HARDVS \cite{hardvs}, DailyAction \cite{dailyaction}, PAF \cite{PAF}, Bullying10K \cite{bully10k}, and UCF101-DVS \cite{hmdbdvs} datasets.

\begin{table*}[!t]
    \centering
    \caption{\it Comparisons with state-of-the-art data-specific methods and zero-shot frameworks on (a-d) object \cite{nimagnet,ncaltech,CIFAR-10-DVS,nmnist} and (e-i) action \cite{hardvs,dailyaction,PAF, bully10k,hmdbdvs} recognition using event camera data. The method categories are distinguished by the `ZS' (\ie, zero-shot) column. All methods are evaluated with the accuracy (\%) metric. For the zero-shot learning framework, we explore its upper bound performance by using optimized text embeddings for the recognition task, denoted by a superscript `$^\text{up}$'.  
    }
    \label{tab:comp}
    \vspace{-.5em}
    \subfloat[All methods are evaluated on the N-ImageNet dataset for object recognition \cite{nimagnet}.
    \label{nimagenet}]{
    \begin{minipage}{0.318\linewidth}
    \setlength{\tabcolsep}{2pt}
    \begin{tabularx}{\linewidth}{lcY}
    \toprule
      Method   & ZS &  Accuracy \\
    \midrule                           
    ECDP \cite{eventpretraing}    & \ding{55} & 68.31 \\
    EventBind \cite{eventbind}  & \ding{55} & 51.40 \\
    MEM \cite{mem} & \ding{55} & 57.89 \\
    \midrule
    EventBind \cite{eventbind} & \ding{51} & 3.05 \\
    EventBind$^\text{up}$ \cite{eventbind} & \ding{51} & 14.45 \\
    ECLIP \cite{eclip} & \ding{51} & 8.72 \\
    ECLIP$^\text{up}$ \cite{eclip} & \ding{51} & 25.05 \\
    EventCLIP \cite{eventclip} & \ding{51} & 4.30 \\
    EventCLIP$^\text{up}$ \cite{eventclip} & \ding{51} & 17.72 \\
    Ours& \ding{51} & 47.84 \\
    Ours$^\text{up}$ & \ding{51} & 57.97 \\
    \bottomrule
    \end{tabularx}
    \end{minipage}
    }
    \hspace{1.5pt}
    \subfloat[All methods are evaluated on the N-Caltech101 dataset \cite{ncaltech} for object recognition. \label{tab:ncaltech}]{
    \begin{minipage}{0.318\linewidth}
    \setlength{\tabcolsep}{2pt}
    \begin{tabularx}{\linewidth}{lYY}
    \toprule
      Method   & ZS &  Accuracy \\
    \midrule                           
    ECDP \cite{eventpretraing}    & \ding{55} & 87.66 \\
    EventBind \cite{eventbind}  & \ding{55} & 94.08 \\
    MEM \cite{mem} & \ding{55} & 90.10 \\
    \midrule
    EventBind\cite{eventbind} & \ding{51} & 67.58\\
    EventBind$^\text{up}$\cite{eventbind} & \ding{51} & 84.11 \\
    ECLIP \cite{eclip} & \ding{51} & 53.88\\
    ECLIP$^\text{up}$ \cite{eclip} & \ding{51} & 74.89 \\
    EventCLIP \cite{eventclip} & \ding{51} & 49.95\\
    EventCLIP$^\text{up}$ \cite{eventclip} & \ding{51} & 72.16 \\
    Ours & \ding{51} & 90.44 \\
    Ours$^\text{up}$ & \ding{51} &93.56 \\
    \bottomrule
    \end{tabularx}
    \end{minipage}
    }
    \hspace{1.5pt}
    \subfloat[All methods are evaluated on the CIFAR-10-DVS dataset \cite{CIFAR-10-DVS} for object recognition. \label{tab:cifar}]{
    \begin{minipage}{0.318\linewidth}
    \setlength{\tabcolsep}{2pt}
    \begin{tabularx}{\linewidth}{lYY}
    \toprule
      Method   & ZS &  Accuracy \\
    \midrule
    ECDP \cite{eventpretraing}    & \ding{55} & 78.00 \\
    DEP \cite{dataefficientpretraining}  & \ding{55} & 78.60 \\
    PSN \cite{psn} & \ding{55} & 85.90\\
    \midrule
    EventCLIP \cite{eventbind} & \ding{51} & 13.36 \\
    EventCLIP$^{\text{l}}$ \cite{eventbind} & \ding{51} & 52.46 \\
    ECLIP \cite{eclip} & \ding{51} & 14.61 \\
    ECLIP$^\text{up}$ \cite{eclip} & \ding{51} & 54.43 \\
    EventCLIP \cite{eventclip} & \ding{51} & 14.05\\
    EventCLIP$^\text{up}$ \cite{eventclip} & \ding{51} & 56.97 \\
    Ours & \ding{51} & 18.48 \\
    Ours$^\text{up}$ & \ding{51} & 60.00 \\
    \bottomrule
    \end{tabularx}
    \end{minipage}
    }

    \subfloat[All methods are evaluated on the N-MINIST dataset \cite{nmnist} for object recognition. \label{tab:nminist}]{
    \begin{minipage}{0.318\linewidth}
    \setlength{\tabcolsep}{2pt}
    \begin{tabularx}{\linewidth}{lYY}
    \toprule
      Method   & ZS &  Accuracy \\
    \midrule
    EventBind \cite{eventbind}    & \ding{55} & 99.27 \\
    E2VID \cite{e2v}  & \ding{55} & 98.30 \\
    RC-GCN \cite{asldvs} & \ding{55} & 99.00\\
    \midrule
    EventBind \cite{eventbind} & \ding{51} & 12.89 \\
    EventBind$^\text{up}$ \cite{eventbind} & \ding{51} & 59.07 \\
    ECLIP \cite{eclip} & \ding{51} &  14.56\\
    ECLIP$^\text{up}$ \cite{eclip} & \ding{51} & 63.37 \\
    EventCLIP \cite{eventclip} & \ding{51} & 11.87 \\
    EventCLIP$^\text{up}$ \cite{eventclip} & \ding{51} & 65.60 \\
    Ours & \ding{51}  & 25.72 \\
    Ours$^\text{up}$ & \ding{51} & 59.60 \\
    \bottomrule
    \end{tabularx}
    \end{minipage}
    }
    \hspace{1.5pt}
    \subfloat[All methods are evaluated on the HARDVS \cite{hardvs} dataset for action recognition. \label{tab:hardvs}]{
    \begin{minipage}{0.318\linewidth}
    \setlength{\tabcolsep}{2pt}
    \begin{tabularx}{\linewidth}{lYY}
    \toprule
      Method   & ZS &  Accuracy \\
    \midrule                           
    ESTF \cite{hardvs}    & \ding{55} & 51.22 \\
    TSM \cite{tsm}  & \ding{55} & 52.63\\
    VideoSwin \cite{videoswin} & \ding{55} & 51.91\\
    \midrule
    EventBind \cite{eventbind} & \ding{51} & 60.66 \\
    EventBind$^\text{up}$ \cite{eventbind} & \ding{51} & 62.53\\
    ECLIP \cite{eclip} & \ding{51} & 37.27 \\
    ECLIP$^\text{up}$ \cite{eclip} & \ding{51} & 39.51 \\
    EventCLIP \cite{eventclip} & \ding{51} & 40.98 \\
    EventCLIP$^\text{up}$ \cite{eventclip} & \ding{51} & 45.44 \\
    Ours & \ding{51} & 64.24 \\
    Ours$^\text{up}$ & \ding{51} & 66.89 \\
    \bottomrule
    \end{tabularx}
    \end{minipage}
    }
    \hspace{1.5pt}
    \subfloat[All methods are evaluated on the DailyAction dataset \cite{dailyaction} for action recognition. \label{tab:dailyaction}]{
    \begin{minipage}{0.318\linewidth}
    \setlength{\tabcolsep}{2pt}
    \begin{tabularx}{\linewidth}{lYY}
    \toprule
      Method   & ZS &  Accuracy \\
    \midrule
    EARSNN \cite{dailyaction}    & \ding{55} & 90.30 \\
    IRSNN \cite{irsnn}  & \ding{55} & 94.60\\
    EJESSR \cite{eventaugpretrain} & \ding{55} & 91.03\\
    \midrule
    EventBind \cite{eventbind} & \ding{51} & 90.33 \\
    EventBind$^\text{up}$ \cite{eventbind} & \ding{51} & 91.17\\
    ECLIP \cite{eclip} & \ding{51} &  85.64 \\
    ECLIP$^\text{up}$ \cite{eclip} & \ding{51} & 86.19 \\
    EventCLIP \cite{eventclip} & \ding{51} &  85.08\\ 
    EventCLIP$^\text{up}$ \cite{eventclip} & \ding{51} & 85.64 \\
    Ours & \ding{51} & 98.62\\
    Ours$^\text{up}$ & \ding{51} & 99.17 \\
    \bottomrule
    \end{tabularx}
    \end{minipage}
    }
    
    \subfloat[All methods are evaluated on the PAF dataset \cite{PAF} for action recognition. \label{tab:paf}]{
    \begin{minipage}{0.318\linewidth}
    \setlength{\tabcolsep}{2pt}
    \begin{tabularx}{\linewidth}{lYY}
    \toprule
      Method   & ZS &  Accuracy \\
    \midrule                           
    EARSNN \cite{dailyaction}    & \ding{55} & 78.10\\
    STCA \cite{stca}  & \ding{55} & 71.20\\
    AEDCM \cite{AEDCM} & \ding{55} & 55.00\\   
    \midrule
    EventBind \cite{eventclip} & \ding{51} & 84.72 \\
    EventBind$^\text{up}$ \cite{eventclip} & \ding{51} & 86.11 \\
    ECLIP \cite{eclip} & \ding{51} & 81.94 \\
    ECLIP$^\text{up}$ \cite{eclip} & \ding{51} & 87.50\\ 
    EventCLIP \cite{eventclip} & \ding{51} & 68.06 \\
    EventCLIP$^\text{up}$ \cite{eventclip} & \ding{51} & 72.22\\
    Ours & \ding{51} & 84.72 \\
    Ours$^\text{up}$ & \ding{51} & 88.89 \\
    \bottomrule
    \end{tabularx}
    \end{minipage}
    }
    \hspace{1.5pt}
    \subfloat[All methods are evaluated on the Bullying10K dataset \cite{bully10k} for action recognition. \label{tab:bullying10k}]{
    \begin{minipage}{0.318\linewidth}
    \setlength{\tabcolsep}{2pt}
    \begin{tabularx}{\linewidth}{lYY}
    \toprule
      Method   & ZS &  Accuracy \\
    \midrule
    R3D \cite{r3d}    & \ding{55} & 66.80\\
    SlowFast \cite{slowfast}  & \ding{55} & 69.00\\
    X3D \cite{x3d} & \ding{55} & 70.80\\
    \midrule
    Eventbind \cite{eventbind} & \ding{51} & 81.80 \\ 
    Eventbind$^\text{up}$ \cite{eventbind} & \ding{51} & 81.92 \\
    ECLIP \cite{eclip} & \ding{51} & 76.26 \\
    ECLIP$^\text{up}$ \cite{eclip} & \ding{51} & 76.45  \\
    EventCLIP \cite{eventclip} & \ding{51} &  73.24 \\
    EventCLIP$^\text{up}$ \cite{eventclip} & \ding{51} & 73.52\\
    Ours & \ding{51} & 87.87 \\
    Ours$^\text{up}$ & \ding{51} & 88.18 \\ 
    \bottomrule
    \end{tabularx}
    \end{minipage}
    }
    \hspace{1.5pt}
    \subfloat[All methods are evaluated on the UCF101-DVS dataset \cite{hmdbdvs} for action recognition. \label{tab:ucf101}]{
    \begin{minipage}{0.318\linewidth}
    \setlength{\tabcolsep}{2pt}
    \begin{tabularx}{\linewidth}{lYY}
    \toprule
      Method   & ZS &  Accuracy \\
    \midrule
    TIM \cite{tim}    & \ding{55} & 63.80\\
    FRMSNN \cite{frmsnn} & \ding{55} & 63.50\\
    3D ResNet \cite{bi3dresnet}  & \ding{55} & 57.90\\
    \midrule
    EventBind \cite{eventbind} & \ding{51} & 79.99 \\
    EventBind$^\text{up}$ \cite{eventbind} & \ding{51} & 81.09 \\ ECLIP \cite{eclip} & \ding{51} &  74.43 \\
    ECLIP$^\text{up}$ \cite{eclip} & \ding{51} & 74.70  \\
    EventCLIP \cite{eventclip} & \ding{51} &  71.62 \\
    EventCLIP$^\text{up}$ \cite{eventclip} & \ding{51} & 72.28\\
    Ours & \ding{51} & 81.32\\
    Ours$^\text{up}$ & \ding{51} & 81.69 \\
    \bottomrule
    \end{tabularx}
    \end{minipage}
    }
\end{table*}

\vspace{-4mm}
\paragraph{Baselines.} We compare with two groups of state-of-the-art methods:
i) zero-shot frameworks for event data on object and action recognition datasets; ii) dataset-specific methods optimized for different datasets.

\vspace{-4mm}
\paragraph{Implementation details.}  We train our method for $50$ epochs with a batch size of 10240, a layer decay of $8.5 \times 10^{-1}$, a gradient clip  of $5$, a learning rate  of $4\times10^{-4}$, weight decay of $5 \times 10^{-2}$, and an AdamW optimizer \cite{adamw}. The learning rate is cosinely scheduled with a warm epoch of 1. We follow \cite{eventpretraing} for augmentations. Our event encoder $f^{\mathsf{evt}}(\cdot)$ uses a ViT-B/16 backbone, initialized from a pre-trained CLIP \cite{evaclip}, and we use the same pre-trained CLIP for the image encoder $f^{\mathsf{img}}(\cdot)$. All the experiments are conducted under the \textit{PyTorch} framework \cite{pytorch}.

\subsection{Object Recognition}

\paragraph{Settings.} For methods belonging to the zero-shot learning framework category (\ie, indicated by a tick in the `ZS' column of \cref{tab:comp}), we explore their upper bound performance by optimizing text embeddings of corresponding class names. The performance of these variations are presented in rows where the method name superscripted with a `$^\text{up}$' symbol. All the zero-shot learning frameworks are based on \texttt{EVA-02-CLIP-B/16} \cite{evaclip} for fair comparisons.

\vspace{-4mm}
\paragraph{Results.} Comparisons with past state-of-the-art methods for object recognition  on the N-ImageNet \cite{nimagnet}, N-Caltech101 \cite{ncaltech}, CIFAR-10-DVS \cite{CIFAR-10-DVS}, N-MINST \cite{nmnist} datasets are given in \cref{tab:comp}~(a-d). Our observations are described as follows:
i) we have the best zero-shot 
performance among the groups of zero-shot learning framework on the four datasets; 
ii) our performance is competitive with the dataset-specific methods on the N-ImageNet \cite{nimagnet} and N-Caltech101 \cite{ncaltech} datasets. However, it falls behind on the CIFAR-10-DVS \cite{CIFAR-10-DVS} and N-MINIST \cite{nmnist} datasets, similar to other zero-shot learning frameworks. This can be attributed to the low-quality 
of these datasets, \eg, they have a small resolution. 
Nonetheless, our methods demonstrate greater robustness than other zero-shot frameworks, achieving higher performance;
iii) ECLIP and EventCLIP exhibit higher performance than our methods when using optimized text embeddings on the N-MINIST dataset. Considering these methods feed event data represented as frames (differing in event frame representations) to CLIP pre-trained on RGB images, it suggests the RGB image pre-trained CLIP remains competitive in extracting discriminative embeddings for events with simple textures.

\subsection{Action Recognition.}

\paragraph{Settings.}  Applying zero-shot learning frameworks to action recognition extends beyond object semantic understanding, requiring the modeling of subtle semantic variations over time.  Besides, the class labels in existing event-based action recognition datasets are often misleading and uninformative, such as  `mobile card 1' and `mobile card 2' from the HARDVS dataset \cite{hardvs}. To address these issues and explore the performance of all zero-shot frameworks, we train a two-layer ViT to temporally aggregate semantics from the visual class embeddings of event frames. This model contains approximately 0.35M parameters and can be trained in 10 minutes. Similarly, the upper bound performance is explored by optimizing text embeddings that are indicated in rows of the method name superscripted with the `$^\text{up}$' symbol.

\vspace{-4mm}
\paragraph{Results.} \cref{tab:comp}~(e-i) compares all methods for action recognition on the HARDVS \cite{hardvs}, DailyAction \cite{dailyaction}, PAF \cite{PAF}, Bullying10K \cite{bully10k}, and UCF101-DVS \cite{hmdbdvs} datasets. Our findings are  threefold:
i) all zero-shot learning frameworks serve as strong semantic extractors. Coupled with the two-layer ViT for reasoning temporal dynamics, they generally achieve better scores than the dataset-specific methods in action recognition. Among them, our method showing the best performance;
ii) consistent with object recognition task, the optimized text embeddings always lead to a better action recognition performance;
iii) our method is significantly better than other zero-shot learning frameworks and dataset-specific methods, on the largest and most complex action recognition dataset, HARDVS \cite{hardvs}, demonstrating the effectiveness of our approach. 

\begin{figure}
    \centering
    \includegraphics{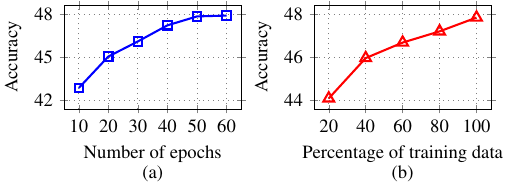}
    \vspace{-2em}
    \caption{\it Ablation study on the number of training epochs and percentage of training data. (a) Training epochs are varied from 10 to 60 with a step size of 10. (b) The percentage (\%) of synthetic data used in training our method is varried from 20\% to 100\% with a step size of 20\%. 
    }
    \label{fig:abl}
\end{figure}

\subsection{Ablations and Discussions}
\label{sec:abl}
Performance of all the ablation studies are evaluated on the largest event-based object recognition dataset (\ie, the N-ImagetNet dataset). 

\vspace{-4mm}
\paragraph{Training epochs.} We study our network convergence by showing the zero-shot object recognition performance variations with respect to the number of training epochs in  \cref{fig:abl}~(a). The training epochs are ranged from 10 to 60, and a step size of 10 is used. A performance saturation is observed after 50 epochs, and motivate us to set the number of training epochs to 50.

\vspace{-4mm}
\paragraph{Data scalability.} We validate the effectiveness of our synthetic event data for training. \cref{fig:abl}~(b) investigates it by varying the percentage of training data from 20\% to 100\% with a step size of 20\%. Consistently, as more synthetic event data is used, the zero-shot object recognition performance is further improved.

\begin{table}[!t]
    \caption{\it Model scalability. We study training with various CLIP image encoder architectures pre-trained from various sources \cite{clip,evaclip}. The performance is evaluated with accuracy (\%). `-' denotes unavailability.}
    \label{tab:parameter_scalability}
    \centering
    \vspace{-.5em}
    \begin{tabularx}{\linewidth}{XXY}
     \toprule
      Model   &  Source  & Accuracy \\
      \midrule
      \multirow{2}{*}{ViT-B/32} & OpenAI \cite{clip} & 32.87 \\
      & EVA \cite{evaclip} & - \\
      \multirow{2}{*}{ViT-B/16} & OpenAI \cite{clip} &  40.47\\
      & EVA \cite{evaclip} & 47.84\\
      \multirow{2}{*}{ViT-L/14} & OpenAI \cite{clip} &  51.44 \\
      & EVA \cite{evaclip} & 57.53 \\
      \multirow{2}{*}{ViT-L/14-336} & OpenAI \cite{clip} & 51.56 \\
      & EVA \cite{evaclip} & 59.17 \\
      \multirow{2}{*}{ViT-bigE/14$^{+}$} & OpenAI \cite{clip} & - \\
      & EVA \cite{evaclip}  & 62.28 \\
      \bottomrule
    \end{tabularx}
\end{table}

\vspace{-4mm}
\paragraph{Model scalability.} \cref{tab:parameter_scalability} studies model scalability in terms of computations and parameters. CLIP based on different image encoder architectures and pre-trained by OpenAI \cite{clip} and EVA \cite{eva} are investigated. Scaling from 86M parameters (ViT-B) to 4.4B parameters (ViT-bigE) for the CLIP image encoder results in a significant increase in zero-shot performance for our methods. Meanwhile, examining the impact of more computations during inference (\eg, ViT-L/14 \vs ViT-L/14-336), we find a positive correlation with zero-shot performance. Jointly with the data scalability explored in the last section, our method demonstrates the appealing scalability in model parameters and training data. 

\begin{figure*}[!t]
    \centering
    \includegraphics{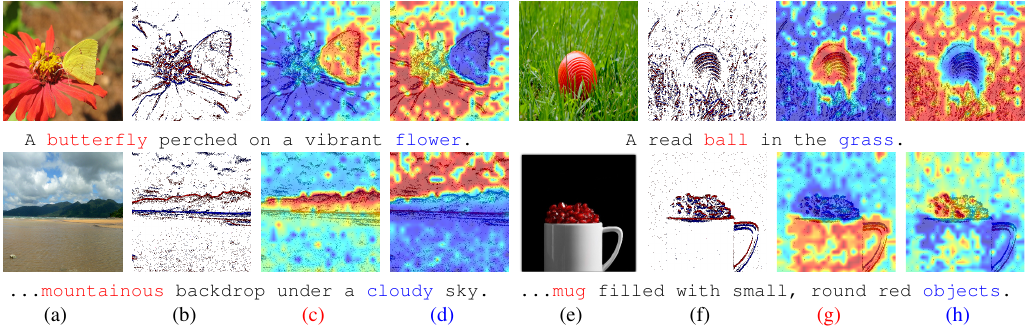}
    \vspace{-2em}
    \caption{\it Samples of heatmaps on event frames. The warmer the color, the more important the area. (a)/(e) are RGB images provided as references. (b)/(f) are event frames. Similarly, red and blue indicate positive and negative events. \textcolor{red}{(c)}/\textcolor{blue}{(d)}/\textcolor{red}{(g)}/\textcolor{blue}{(h)} are heatmaps with respect to texts highlighted in the same color, where the text descriptions of the image are provided in the bottom of each row.
     }
    \label{fig:heatmap}
\end{figure*}

\begin{figure*}[!t]
    \centering
    \vspace{-1em}
    \includegraphics{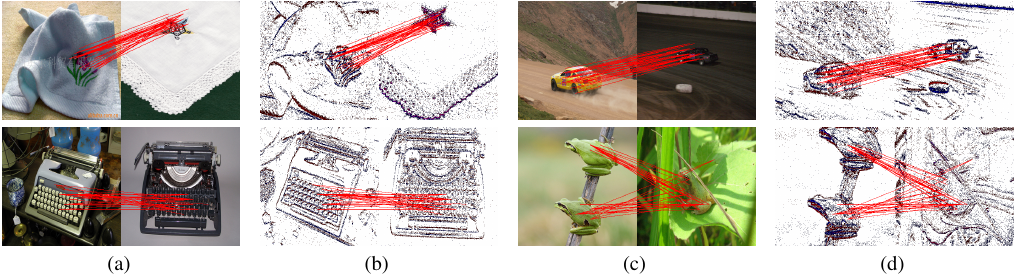}  
    \vspace{-1em}
    \caption{\it Samples of cross correspondences between two semantically similar event frames. The corresponding RGB images of (b)/(d) are shown in (a)/(c) as references. Note that there are some spatial misalignments between the RGB and event frames. The top cross correspondences extracted from the event frames in (b) and (d) are visualized as red lines.}
    \label{tab:cor}
\end{figure*}

\begin{table}
\vspace{-1em}
\centering  
\setlength{\tabcolsep}{3pt}
\caption{\it Example  
of integrating our event encoder in place of the one used by LLaVA \cite{llava}. The corresponding RGB image of the event frame is shown on the right for reference.}
\label{tab:visual_example}  
\vspace{-.5em}
\small
\begin{tabularx}{\linewidth}{lX}
\toprule
\multicolumn{2}{l}{\bf Example conversation}  \\
\midrule
&  \includegraphics[width=.305\linewidth, height=.305\linewidth]{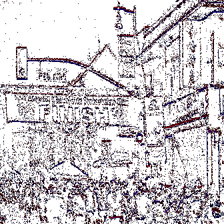} \quad \quad \quad \includegraphics[width=.305\linewidth,height=.305\linewidth]{fig/conversation_event.png} \\
User: & Caption the image in 80 words. \\
Assistant: & \parbox[t]{\linewidth}{A bustling city street scene with a crowd of people, a mix of individuals and groups, some walking, others standing, all engaged in their own activities. The atmosphere is vibrant and dynamic, with a sense of movement and energy. The architecture and signage suggest a dense urban environment, possibly in a downtown or central district. The image captures the diversity and liveliness of city life.} \\
\bottomrule
\end{tabularx}
\end{table}

\vspace{-4mm}
\paragraph{Event and text data.} The reader may wonder about the performance of our methods when using synthetic event data and text data. We fine-tune our network using the synthetic event data on the CC3M dataset \cite{cc3m}. However, we observe a decrease in zero-shot performance during training, with an accuracy of only 14.84\%, which is significantly lower than the 47.84\% accuracy of our method. The performance decrease suggests the effectiveness of our approach.

\vspace{-4mm}
\paragraph{Model analysis.} We evaluate our model from three perspectives:  i)  \cref{tab:visual_example} demonstrates conversations with a vision-language model, where our event encoder is integrated as a replacement for the CLIP image encoder used in LLaVA \cite{llava}; ii) \cref{fig:heatmap} visualizes heatmaps of event data in relation to corresponding texts; iii) \cref{tab:cor} extracts the sparse cross-correspondences between two semantically similar event frames.  Overall, our model exhibits a robust ability to understand semantic content. For instance, in \cref{fig:heatmap}, the model trained on event and RGB data pairs effectively captures meaningful textual semantics. Specifically, on the right side of the second row, the model accurately identifies the last word, `objects', described as `small' and `round', as referring to pomegranates. This strong semantic comprehension ability underpins our performance in tasks such as object and action recognition.

\vspace{-4mm}
\paragraph{Limitations.} 
While progressing in zero-shot object recognition, a gap remains in developing a generalist foundation model for event camera data. However, as shown by \cite{llava,instructblip}, a versatile pre-trained CLIP image encoder is a key component for such models, which this paper investigates.

\vspace{-4mm}
\paragraph{More results.} Refer to more ablations, visualizations, and experiments in our supplementary material.

\section{Conclusion and Broader Impact}
We propose a framework to unlock zero-shot object recognition performance within the event camera data community. Our method is trained with event and RGB data pairs. To address the scarcity of these data pairs, we study a pipeline to synthesize event data from still RGB images. By analyzing the limitations 
of existing approaches, our key insights are to perform the 
scalar-wise semantic modulation during training. Experiments on standard object and action recognition benchmark datasets extensively demonstrate the superiority of our approach compared to previous state-of-the-art methods.

\vspace{-4mm}
\paragraph{Broader impact.} 
We believe our method is promising to be applied in training a generative event-vision-language foundation model. We hope it inspires future work.

{
    \small
    \bibliographystyle{ieeenat_fullname}
    \bibliography{main}
}

\end{document}